# A Hybrid Framework for Real-Time Data Drift and Anomaly Identification Using Hierarchical Temporal Memory and Statistical Tests

**Subhadip Bandyopadhyay**
Global AI Accelerator,
Ericsson, Bangalore, Karnataka, India.
E-mail: subhadip.bandyopadhyay@ericsson.com

**Joy Bose**
Global AI Accelerator,
Ericsson, Bangalore, Karnataka, India.
E-mail: joy.bose@ericsson.com

**Sujoy Roy Chowdhury**
Global AI Accelerator,
Ericsson, Bangalore, Karnataka, India.
E-mail: sujoy.roychowdhury@ericsson.com

*Corresponding author*: Subhadip Bandyopadhyay

**Abstract**

Data Drift is the phenomenon where the generating model behind the data changes over time. Due to data drift, any model built on the past training data becomes less relevant and inaccurate over time. Thus, detecting and controlling for data drift is critical in machine learning models. Hierarchical Temporal Memory (HTM) is a machine learning model developed by Jeff Hawkins, inspired by how the human brain processes information. It is a biologically inspired model of memory that is similar in structure to the neocortex, and whose performance is claimed to be comparable to state of the art models in detecting anomalies in time series data. Another unique benefit of HTMs is its independence from training and testing cycle; all the learning takes place online with streaming data and no separate training and testing cycle is required. In sequential learning paradigm, Sequential Probability Ratio Test (SPRT) offers some unique benefit for online learning and inference. This paper proposes a novel hybrid framework combining HTM and SPRT for real-time data drift detection and anomaly identification. Unlike existing data drift methods, our approach eliminates frequent retraining and ensures low false positive rates. HTMs currently work with one dimensional or univariate data. In a second study, we also propose an application of HTM in multidimensional supervised scenario for anomaly detection by combining the outputs of multiple HTM columns, one for each dimension of the data, through a neural network. Experimental evaluations demonstrate that the proposed method outperforms conventional drift detection techniques like the Kolmogorov-Smirnov (KS) test, Wasserstein distance, and Population Stability Index (PSI) in terms of accuracy, adaptability, and computational efficiency. Our experiments also provide insights into optimizing hyperparameters for real-time deployment in domains such as Telecom.

**Keywords-** Hierarchical Temporal Memory (HTM), Sequential Probability Ratio Test (SPRT), time series, real-time anomaly detection, data drift detection, streaming data analysis, hybrid machine learning models, telecom network monitoring, AI powered data drift detection

## 1. Introduction

Data drift is a phenomenon where the generating model of some data changes over time. In presence of data drift, any model trained on older data becomes less relevant and its performance degrades in context to the recently available data. If a machine learning (ML) model is constructed based on some training data, it is necessary to periodically retrain the ML model based on recent data to maintain a consistent performance as the training data distribution may vary over time in presence of data drift.

Data drift detection is an age-old problem (Kadam, 2019, Wadewale et al., 2015, Gemaque et al., 2020, Wang & Abraham, 2015). It has historically appeared in various formats in Statistical Science literature including change point detection, Statistical Process Control or statistical inference of two and/or multiple sample tests for population homogeneity. Recently, with the explosion of Artificial Neural Network based approaches, applications of Deep Learning techniques for data drift detection have notable presence in the current literature. However, for time sensitive applications, current emphasis on the near real time performance of data drift detection methodologies have got more attention, e.g., in various use cases in the Telecom and Finance domain.

Some examples of applications where data drift or anomaly detection is useful include fraud detection in banking and finance, by detecting anomalies in transactions, in healthcare industry by identifying anomalies in vital signs of patients, in Internet of Things (IoT), by detecting sensor anomalies, and in cybersecurity by detecting network intrusions. There are also recent real-life applications from various domains where neural networks have been used to detect anomalies and other patterns in real life streaming data, such as in the case of predicting nanofluid flows (Yaseen et al., 2023, Rawat et al., 2023, Mishra et al., 2023, Bhadauria et al., 2024).

Hierarchical Temporal Memory (HTM) (Hawkins et al., 2019, Numenta whitepaper, 2019) is a model of neural networks that is more biologically plausible than deep learning. It is an unsupervised method, with no separation between train and test stages. Consequently, it performs well with streaming data. The network architecture in HTMs closely mirrors the cortical columns of the neocortex of the mammalian brain. The nature of HTM methodology falls under the sequential learning paradigm which is very much relevant in use cases where time series data is involved.

Sequential learning is a rich and developed field in Statistics which is currently being applied in multiple other domains for its due relevance and generality (Sutskever et al., 2014). Sequential Probability Ratio Test (SPRT) (Piegorsch et. al., 2011) is one such basic tool used primarily for sequential statistical inference. From this context, though a confluence of sequential learning approaches and HTM based techniques is apparently very much natural, however in the existing literature we hardly observe any conscious effort of such a symbiosis. We consider un-supervised scenario for data drift detection which is more realistic scenario as obtaining labelled data under drifted scenario is often not possible practically as identification of drift is needed at the outset. HTM and SPRT have been applied in different use cases involving time series data like time series anomaly detection, Statistical Process Control and forecasting, to name a few. Specific applications can be found in Wu et al. (2018) for HTM and in Schoonewelle et al. (1995) for SPRT.

In this paper, we explore building an online solution for data drift detection (in un-supervised scenario) and anomaly detection (in supervised scenario) by combining HTM and SPRT. The idea of combining these two sparsely existing approaches is driven by the need of an online anomaly and data drift detector with minimum or no re-training load. Since HTM does not need re-training and HTM and SPRT both are suitable for near real time adoption, an intuitive solution is to combine both. The advantage of this approach is that no frequent retraining of drift detection algorithm is needed for a near-real time application.

It can be noticed that HTM and SPRT serves two different purposes. Where HTM basically predicts the similarity of the next time point data with most recent data, SPRT compares the current data pattern with the historical data mostly under some assumption of the data distribution. In a real-life scenario, the distribution of the data is not known and we bypass this problem by leveraging HTM output and formulating it as a binary decision, namely, if the current data point is coming from historical data distribution. With this formulation, irrespective of the real data distribution, the reformulated sequence of data can be modelled under sequence of Bernoulli random variable since the current data point is either from the historical distribution or not from the same distribution. Hence the rich field of SPRT can be applied subsequently to infer a data drift, namely, if the current data point is not from the historical data distribution then a drift can be detected to be in force.

There are existing work on concept drift detection on streaming data (Souza et. al, 2018, Pesaranghader et. al., 2018). There are existing techniques for testing homogeneity of population based on two samples, e.g., Kolmogorov-Smirnov (KS) test, Wasserstein distance, Population Stability Index or Jensen-Shannon Divergence (Thomas et. al., (2006), Endres et. al.(2003), Yurdakul B. (2018) ) which are suitable for a rolling window based formulation for online data drift detection but many of them suffers high false positive rate as we illustrate in our simulation experiments in section 4.

In un-supervised scenario, the proposed approach is divided into two phases. In the first phase, HTM is utilized to continuously model the probability of similarity of incoming sample data with respect to observed data points from recent past. The first phase output is consumed in the second phase to built an SPRT based drift detection algorithm.

In supervised scenario, we assume that either knowledge from SMEs or any other source is available for outlier labelling. HTM layer, as in supervised scenario, is considered for each data dimension. We propose a HTM output combiner via Neural Net (NN) construction for anomaly detection. We have used a combination of existing models to generate the ground truth for the supervised scenario as experts' data labelling was not available readily.

The key contributions of this work are:
   (a) A hybrid methodology combining HTM and SPRT for efficient drift detection in streaming data.
   (b) Extension of HTM for multivariate anomaly detection using a neural network combiner.
   (c) An evaluation of the proposed framework against other methods, highlighting its better performance and adaptability.
   (d) Insights into hyperparameter tuning for optimizing the framework's performance.

The rest of the paper is formatted as follows. Section 2 reviews related work in the area of HTMs, SPRT and data drift detection. Section 3 introduces the motivation and describes the application of combined HTM and SPRT for drift detection for the unsupervised scenarios. Section 4 extends the approach of combining HTM using a neural network combiner to multivariate and supervised scenario. Section 5 concludes the paper with discussion on limitation and future directions.

**2. A brief introduction to HTM and related work**
In this section, we introduce HTMs and its application in anomaly detection.

## 2.1 Hierarchical temporal memory (HTM)

Hierarchical temporal memories were first introduced by Jeff Hawkins, based on the groundwork in the book "on intelligence" and subsequent work done in Numenta (Hawkins et al., 2019, Numenta whitepaper, 2019), the company he founded. HTMs are biologically inspired memory models that resemble the structure of the neocortex and function as associative memory neural networks. They are more biologically plausible than common deep learning models such as CNNs.

The network architecture in HTMs closely mirrors the cortical columns of the neocortex of the mammalian brain. The structure of the HTM replicates properties of brain's neocortex including hierarchical structure, sparse distribution, Hebbian learning (following Hebb's law), noise tolerance and adjustment to changing data or data drift. HTMs are trained in an unsupervised way, with no separation between train and test stages. Consequently, they are suited to work well with streaming data.

Computational aspects of HTMs include encoding input data as binary sparse vectors (SDRs), structuring associative memory, enabling online and continuous learning, predicting the next output for each input, and generating an anomaly score that estimates the likelihood of the current input being anomalous.

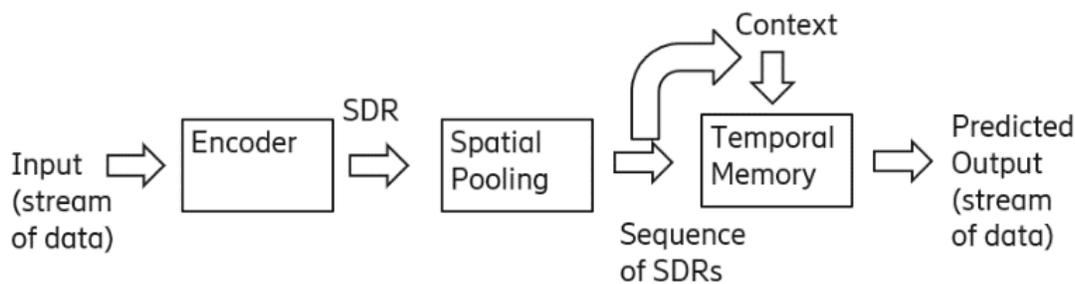

**Figure 1.** High level architecture of HTMs

HTMs have the following components:

- **Encoder**: this is a layer of neurons that takes the input stream of data and encodes it as a binary vector. All types of input data, whether it be numbers, text, images, audio or video, all are encoded as binary vectors.

    $I_t$ = input data at time t

    Output of encoder = $x_t$ = Binary representation of $I_t$  [1 0 0 1 0 0 ….. 1]

- **Spatial Pooler**: This is a layer that converts the binary vector using an encoding scheme called sparse distributed representation or SDR. Sparse refers to the fact that most of the bits are 0s and a few (around 2%) are 1s. Multiple neighboring HTM cells are activated for each bit, and the overlap between two SDRs is the measure of similarity. It takes the binary vectors as input and activates (sets to a predictive state) columns which have the highest overlap with the input vector.

    Output of spatial pooler = $a(x_t)$ = SDR representing $x_t$

- **Temporal memory**: This layer incorporates the temporal aspects of the data, i.e., it remembers the relation between the current data and previous data, by having an explicit memory to represent the state or context. All the data is encoded as SDRs as mentioned earlier. For each new data that comes in, it predicts what the next data should be. It implements Hebbian learning (Hebb, 2005) which is a biologically plausible learning algorithm that is based on Hebb's rule: cells that fire together, wire together. Since the SDRs are binary vectors, only some bits are 1s, which activate the HTM neurons connected to those bits and set the connection weight to 1.

Output of temporal memory = $\pi(x_t)$ = SDR predicting $a(x_{t+1})$ where $x_{t+1}$ means the input at time (t+1)

- **Anomaly Detector**: This module outputs the log likelihood of the current input being an anomaly. It takes two inputs: the SDR predicting $a(x_{t+1})$ and the output of the spatial pooler $a(x_t)$ and outputs the log likelihood of anomaly at time t.

  Output of anomaly detector = $l_t$ = Log likelihood of anomaly

Figure 1 shows the high-level architecture of the HTMs.

There are a number of available libraries that implement HTMs, such as Numenta, which is proprietary and HTM-core which is open-source. In this paper, we have used the open-source htm-core (Github, 2019) library to implement HTMs.

## 2.2 Anomaly detection aspect of HTMs

The basic principle of using HTMs for anomaly detection is as follows: the HTM learns a sequence of data and predicts what the next data should be at any given time, based on what it has learnt in the past. If the prediction does not match the actual new data that is coming in, that means a potential anomaly.

The system for anomaly detection thus compares the predicted data at time t with the actual data, and gets a prediction error, from which it decides the likelihood of an anomaly. It can work in real time without any separation of test and training data.
There have been a few studies in the literature (Wu et al., 2018, Numenta Whitepaper, 2019, Barua et al., 2020, Ahmad & Purdy, 2016, Anandharaj & Sivakumar, 2019) that show how HTMs perform comparable or better for anomaly detection than other conventional ML techniques. Moreover, HTMs have the added advantage of one-shot learning (only one instance of a data is sufficient for the HTM to learn the associations), which makes them suitable for use with streaming data.

Figure 2 shows the overview of an anomaly detection system using HTMs.

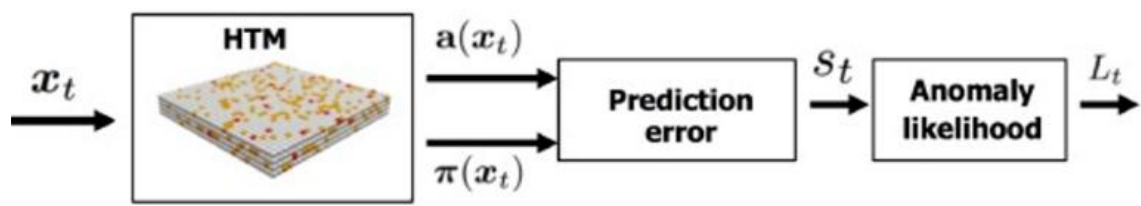

**Figure 2.** Overview of architecture for anomaly detection system using HTMs

Figure 3 shows the architecture of a single HTM for anomaly detection, where the input stream is in a single dimension, with an additional module called drift detector that detects the drift in the input stream, if any.

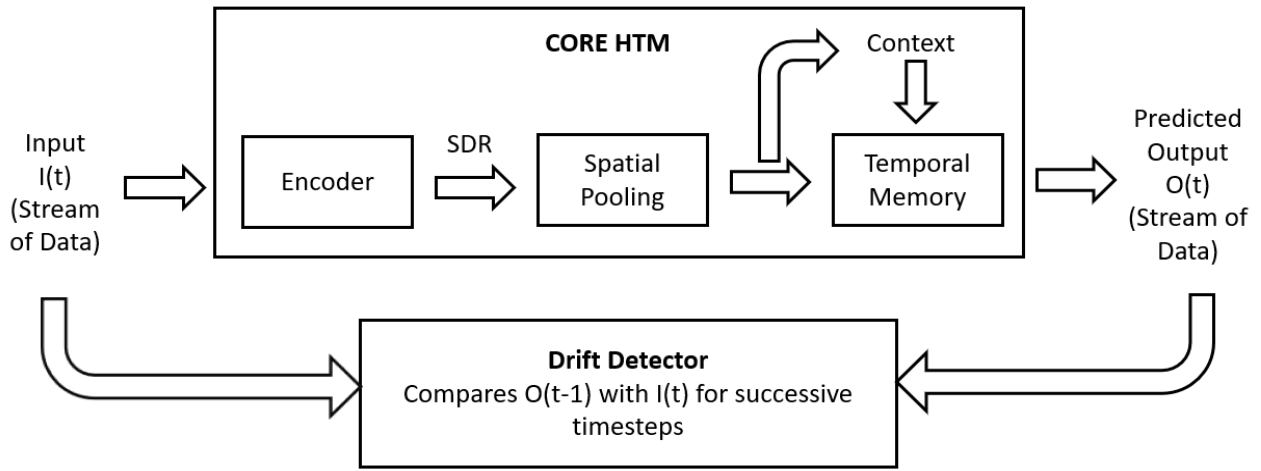

**Figure 3.** Architecture of using HTM for anomaly detection in a single dimensional scenario

Figure 4 shows the architecture of the anomaly detector where the input stream is in multi-dimensional.

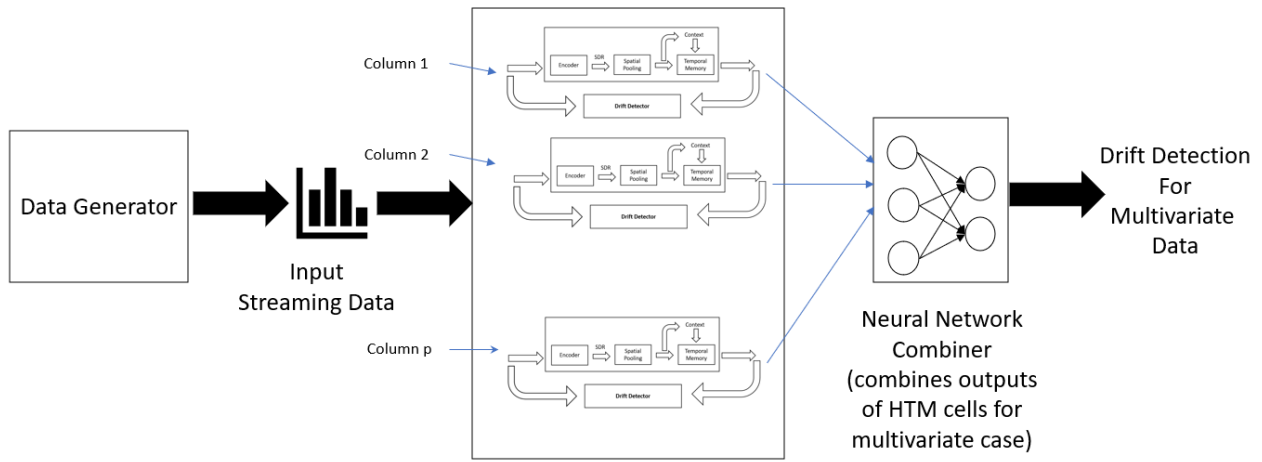

**Figure 4.** Architecture of using HTM for anomaly detection in a multi-dimensional scenario

## 3. Data drift detection through hybridization of HTM and SPRT in unsupervised scenario

### 3.1 Motivation and overview of the solution:

Data drift is a critical issue that is needed to be addressed to make any data based algorithmic task, as a part of any application pipeline, consistent and reliable in field performance (Gemaque et al., 2020). A generic problem for any data-based drift detection algorithm is the frequent need of retraining. Each time a drift is detected, the underlying data distribution changes and hence re-calibration of algorithm is needed with respect to the new data distribution. For non-parametric approaches like Chebyshev's inequality driven k-σ limit-based drift detection or z-statistic based drift detection, the mean and variance re-computation is necessary when the data distribution changes after a data drift.

Any automated pipeline must address this critical requirement through re-training, which ultimately increases the cost and complexity of maintaining an online, real-time, or near-real-time autonomous drift detector.

To address this problem, we propose to break this requirement into two hierarchical parts. In the first phase of our approach, a HTM layer is utilized to continuously model and output the probability of similarity of incoming sample data with the observed data points from recent past. HTM, by its formulation, continuously adapts to changes in the recent data distribution within a reasonable time lag. Hence the probability output from the HTM layer consistently re-adjusts itself with respect to new environment, also known as the drift in the data distribution.

The first stage outputs feed an SPRT based drift detection algorithm in the second stage. We re-formulate the problem, through formulation of appropriate hypothesis, into a Bernoulli distribution based SPRT for Bernoulli parameter.

The SPRT in the second stage can continuously operate within the lagged time window to assess the similarity of incoming data with recent past observations and detect drift. When a drift is detected, after the lagged time, HTM re-adjust itself with the change in data distribution. Subsequently, the SPRT module continues the drift detection seamlessly, now with respect to the new data without any break. Thus, the whole flow continues seamlessly.

### 3.2 Re-Formulation of the problem

We set the assumptions, definitions and notations below before formal algorithmic steps.

**Assumption 1:** We assume that the 'd' most important dimensions are identified through exploratory data analysis.

For a specific dimension, we re-formulate the problem of data drift detection in the following way.

Define $p_i$ = Probability that the $i^{th}$ dimension follows a new distribution due to data drift.

It can be noted that $p_i$ is estimated and outputted by the HTM layer consistently corresponding to the current data point as a quantification of the chance that the current observation is significantly different than the recent past data pattern. Hence a sequence of $p_i$ is obtained consistently from the HTM layer.

**Proposition 1:** Following the definition of $p_i$ stated above and a suitable $p_{null}$ and $p_{alt}$ in (0,1), a data drift can be associated with the hypothesis $p_i \geq p_{alt}$, which represent that the data pattern has changed significantly (e.g., for a $p_{alt} = 0.65$). On the other hand, the hypothesis $p_i \leq p_{null}$ (e.g., $p_{null} = 0.45$) can be associated with the hypothesis that there is no drift. In the similar line of thought, the hypothesis $p_{null} < p_i < p_{alt}$ corresponds to the lack of data support scenario to decide that a data drift has happened and hence an inference on data drift occurrence needs more data points. Thus a data drift detection problem can be formulated as a sequential hypothesis testing problem and SPRT is a natural candidate as the apt tool for testing data drift.

**Proposition 2:** $p_{alt}$ and $p_{null}$ can be looked upon as two tuning parameter which can either be obtained as users' input depending on the conservativeness of user's perception, or based on SME knowledge or set suitably based on historical data driven analysis in the specific use case context.

Note that setting the value of $p_{null}$ below 0.5 safeguards higher chance of a false positive in drift detection.

Next we re-formulate HTM layer output as Bernoulli random variables distribution through a re-parametrization.

**Definition 1:**
Define $htm_t$ = output of the HTM (for a specific data dimension) at time point t. This represents the likelihood that the new observation at time t deviates from historical data.

**Definition 2:**
Let us define $c_t$ = 1 if $htm_t$ > bin_threshold where bin_threshold in range (0,1)
         = 0 otherwise,

**Lemma 1:** Under the assumption of independent sample data, $c_t$ ~ Bernoulli(0, $p_i^*$), t=1,2,.. where $p_i^*$ represent probability of data drift event, for the i-th data dimension, corresponding to the data drift hypothesis proposed in Proposition 1.

*Proof:* The proof follows directly from the definition of Bernoulli distribution once we note the following points.
  1. $p_{alt}$ mentioned in the hypothesis in Proposition 1 and bin_threshold in the definition of $c_t$ has one to one mapping.
  2. The event $c_t$ = 1 and onset of data drift hypothesis in Proposition 1 corresponds to each other.
  3. $P(c_t = 1) = p_i^*$.
  4. The sequence { $c_t$, t=1,2,..} are sequence of independent events.

**Remark 1:** Implication of Lemma 1 is that the random event, 'i-th data dimension's distribution remained unchanged at time point t , t=1,2,..', can be considered approximately following Bernoulli ($p_i^*$) where $p_i^*$ quantifies the similarity likelihood of t-th observation and recent past data points.  Note that under no drift hypothesis, $p_i^* \leq p_{null}$, for i = 1,..,d.

**Remark 2:** We can extend Proposition 1 for each of the data dimension and conceptualize $p_{null}$ and $p_{alt}$ varying accordingly. This generic approach can accommodate more flexibility in the dimension wise drift detection where the definition of a drift can be tuned as per the corresponding data dimensional feature and importance. For more important data dimensions (typically determined by use case or an EDA based on historical data) we can adapt conservative view to reduce false discovery rate.

**3.3 Constructing SPRT for checking data drift in a single dimension**

First, we stet the relevant assumptions and definitions.

**Assumption 2:** We assume that the observed data has a time stamp so that we can perceive it as a data stream.

For a single data dimension, we want to derive a SPRT pipeline based on the HTM output to test, in continuum, the null hypothesis that there is no data drift against the alternative that a data drift has happened.

To construct a SPRT, we need upper threshold of two types of error, namely type 1 and type 2 error.

**Definition 3:**
  (i)   Let us define a = probability of type 1 error = P (we infer drift has happened whereas it has not).
  (ii)  Define b = probability of type 2 error = P (we infer no drift where actually there is a drift)

It can be re-emphasized that these choices should be governed by specific business use case. Where retraining is less costly, we can relax a and increase recall.

In our simulation experiment, we have considered a = 0.05 and b = 0.005 that gave reasonably good detection.

**Remark 3:** We introduce a rescaling of $htm_t$ and work with the rescaled values to construct $c_t$ (ref. Definition 1 in previous subsection). The re-scaled score from HTM encodes the dissimilarity between the observation at time point t and the prediction of the observation from HTM pipeline in a same way like original HTM output. This re-scaling is a window based smoothing type approach to regularize the abrupt fluctuation that may result into computational instability.

We describe this re-scaling process in the next subsection 3.4.

Following Lemma 1 in previous subsection, we can consider the sequence $\{c_t ; t=1,2,..\}$ is coming from a Bernoulli distribution with parameter 'p' where p quantified the probability that the observation at time point 't' is a sample from a new distribution. Thus testing for a low value of 'p' will lead to test for a data drift as stated in Theorem 1 below.

**Definition 4:**
Define $Cm_t$ = cumulative sum of $c_t$ up to time point t.

**Theorem 1:**
Following the SPRT rule for Bernoulli sequence $\{c_t\}$, we identify a drift has set in if:

$$Cm_t > \text{Upper\_limit} = \frac{\log(\frac{1-b}{a}) + t*[\log(\frac{1-\text{p\_nul}}{1-\text{p\_alt}})]}{\log(\frac{\text{p\_alt}}{\text{p\_null}}) - \log(\frac{1-\text{p\_alt}}{1-\text{p\_nul}})} \quad (1)$$

We identify no change of data distribution if :

$$Cm_t < Lower_{\_limit} = \frac{\log(\frac{b}{1-a}) + t*[\log(\frac{1-p\_nul}{1-p\_alt})]}{\log(\frac{p\_alt}{p\_null}) - \log(\frac{1-p\_alt}{1-p\_nul})} \quad (2)$$

*Proof:* Following Proposition 1,2 and Lemma 1, since we have framed the data drift detection problem as testing of hypothesis based on Bernoulli formulation of the HTM output data SPRT construction, the proof follows directly from the theory of SPRT (e.g., Piegorsch & Padgett, 2011). More explanation around equations (1) and (2) can be explored in Piegorsch & Padgett, 2011.

**Remark 4:** For the multidimensional scenario, SPRT based control chart can be run simultaneously for all the dimensions and a final data drift detection rule can be flexibly formulated combining dimension wise drift detection inference. For example, a conservative drift detection rule can be formulated as, if drift is detected for at least one of dimensions, a data drift is inferred as final output. This additional layer gives us a flexibility of use case specific drift detection formulation.

We continue monitoring the process and identify onset of a data drift when the process is out of control or no data drift when the process is within control. Whenever a data drift onset decision is reached, the SPRT is started afresh from the next time point. Note that the hyperparameters a,b, $p_{alt}$ and $p_{null}$ remains unaltered.

The advantage that HTM brings is that when the drift is completed, the estimated HTM probability will stabilize in lower range like (0,0.5) automatically and hence the process will again show as under control, or as in our case, no data drift as all the new data is coming from the new distribution after a drift.
We will be able to identify the drift window directly from observing the charts ( e.g., Figures 5 – 7 in univariate scenario).

A data drift window begins when at least one data dimension exceeds the control threshold based on $Cm_t$ comparison.

**Remark 5:** After a drift starts, monitoring each data dimension continues with SPRT being re-started and we can identify when the drift ends by identifying time point when a new drift is detected in any of the single dimension. During this time duration, SPRT check for all the dimensions remain below lower threshold as the data points in this duration is generated by the new data distributions.

It can be noted that a more appropriate way to accommodate a data drift detection in a multidimensional scenario is to consider both HTM and SPRT in multidimension. Research in multidimensional HTM is yet to be addressed in literature. Therefore, we consider unidimensional HTM output. Though multidimensional SPRT is well addressed in literature, we however considered a rule-based approach on unidimensional SPRT to illustrate the concept.

**3.4 Rescaling of HTM output for anomaly score generation in unsupervised scenario**

The output of core HTM is a similarity score of the observation given all history. From our simulation experiment, we observed that working with direct HTM output is not suitable for creation of the binary sequence $\{c_t, t=1,2,..\}$.
We create an anomaly score with the same in the SPRT phase discussed as follows.

Let at time $t_0$, we have run HTM and we have output from HTM core as htm_value_$t_0$ . Let corresponding observed value is obs_val_t.

Now consider a historical window of size $w_{sz}$. This window size is again a user driven or a data driven input which will depend on the business use case, SME and/or historic data.

We compute rolling standard deviation $\sigma_{roll}$ from this data.

For the t-th observation, anomaly score is defined as:

$$\text{anoml\_score} = \frac{\text{absolute(htm\_value\_t} - \text{obs\_val\_t)}}{k * \sigma\_roll} \quad (3)$$

Thus the final HTM output is obtained from anoml_score as follows:

$$\text{htm}_t = \text{anoml\_score if anoml\_score} < 1$$
$$= 1 \text{ if new\_anoml\_score} > 1. \quad (4)$$

We do this for all the data points in the time series and create the sequence { $c_t$ , t=1,2..} as Definition 2.

### 3.5 Algorithmic steps for data drift detection

Consolidating the discussion in section 3.1 and 3.2, we formally state the algorithm steps below.

**Algorithmic steps for data drift identification**
Input: Data on relevant dimension streamed as input, L (window length for htm output regularization), a, b, $p_{null}$ , $p_{alt,}$ bin_threshold (to construct $c_t$), anomaly limit k
Output :    Starting time point of a possible data drift

Execute the following step for the incoming data stream in continuum.
1. Input data stream to HTM layer and collect output $htm_t$ for t=1,2,.. upto current time point.
2. Re-scale HTM output as per equation (3) and (4) in section 3.4.
3. Compute $c_t$ following Definition 2 and $Cm_t$ following Definition 4.
4. Compute Upper$_{limit}$ and Lower$_{limit}$ following equation (1) and (2) in Theorem 1.
5. Compare $Cm_t$ with Upper$_{limit}$ and Lower$_{limit}$ to either detect data drift or continue monitoring as per Theorem 1.

### 3.6 Experiment on drift detection with simulated data in unsupervised scenario

To illustrate the concept, we consider univariate scenario where data is generated from a normal distribution and the drift is present as dynamically changing mean over time. We consider three basic pattern of data drift, namely, periodically change, slowly but monotonical change and abrupt changing in the mean.

After running the drift detection algorithm, vertical lines in the data plot represent onset of a drift phase. The parameter sets corresponding to these experiments is mentioned in table 1 later.

*Periodically time varying mean:* The data is drawn from a normal distribution with a mean function with period 250 and amplitude 0.5.

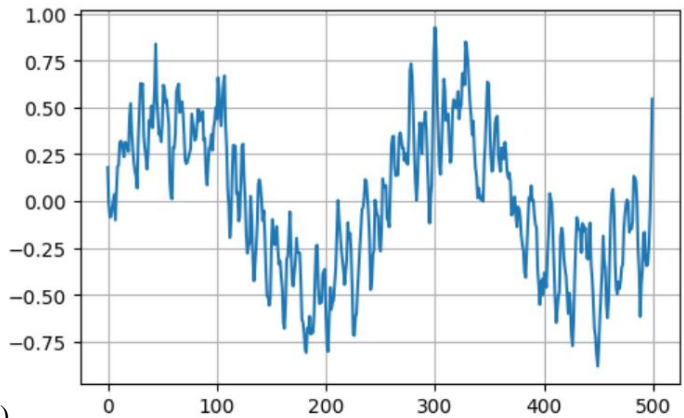
(a)

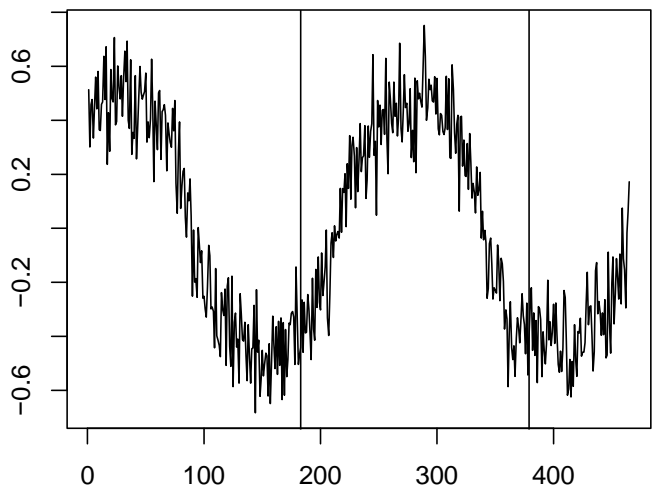
(b)

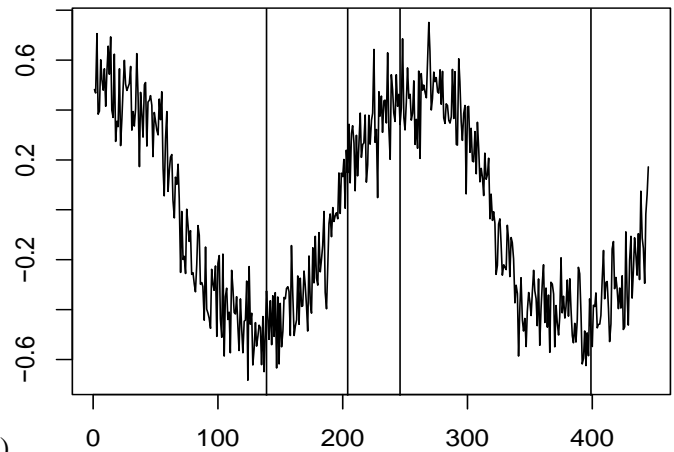
(c)

**Figure 5.** Time (X-axis) – Vs – value (Y-axis) Plot (a) Gaussian with periodically changing mean (b) Drift start time (vertical lines) with historical window size 25 (c) Drift start time(vertical lines) with historical window size 10

*Monotonically time varying mean:* In the below example, samples are generated from a normal distribution with monotonically changing mean over time. The changing mean pattern is determined by a 3-rd degree polynomial with randomly generated coefficient from a standard normal distribution.

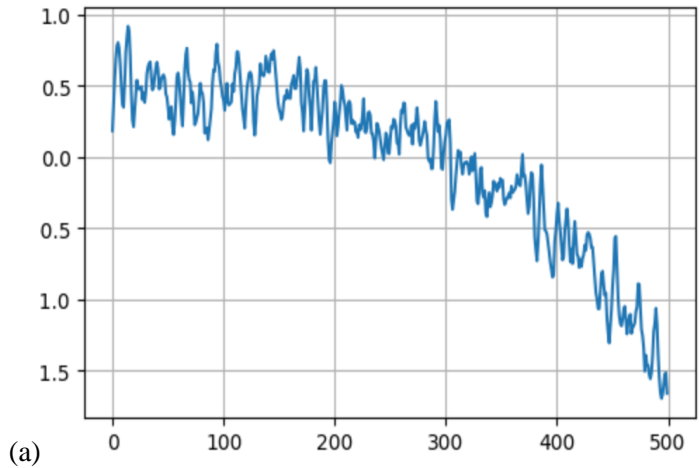

(a)

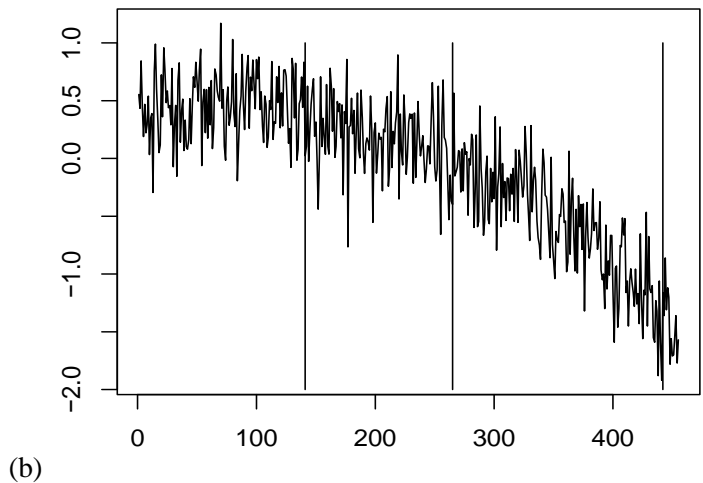

(b)

**Figure 6.** Time(X-axis) – Vs – value(Y-axis) plot (a) Plot of rescaled data with monotonically changing mean (b) Detected drift start time point (Vertical lines)

*Abrupt mean shift:* The third pattern considered is an abrupt data drift through a one-time increase in the gaussian mean by 2 units. This change is imparted at time point t = 250 and we can notice the distinctness in the data distribution before and after t=250.

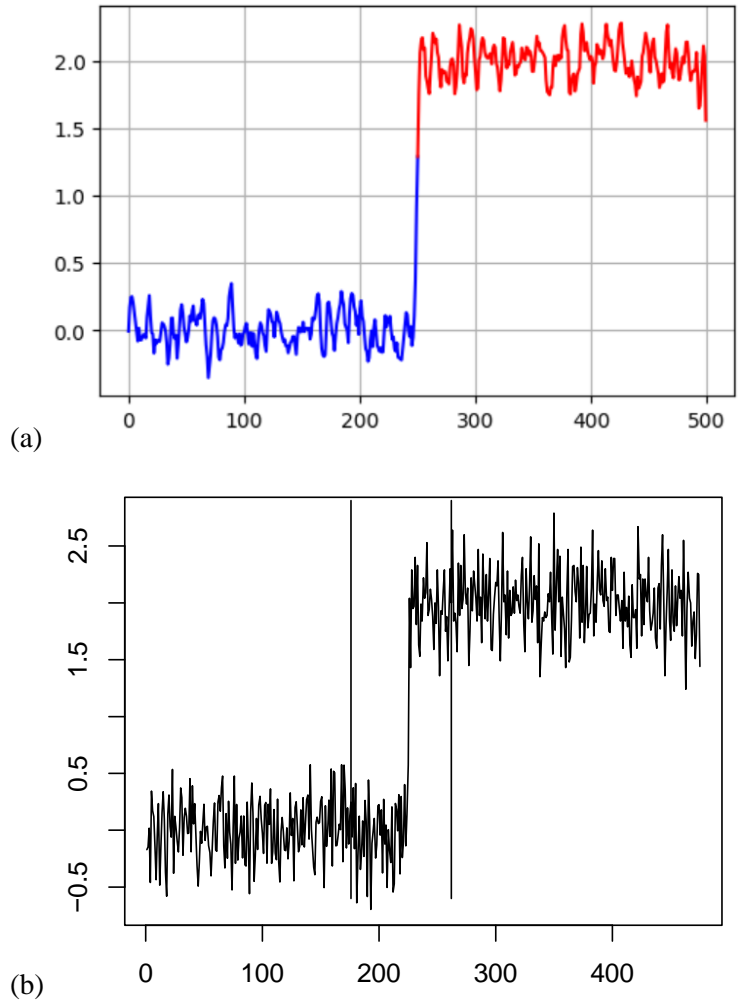

(a)

(b)

**Figure 7.** Time(X-axis) – Vs – value(Y-axis) plot (a) Plot of rescaled data with abrupt mean change (b) Detected drift start time point (Vertical lines)

**3.7 Discussion of the experimental result of the proposed approach**

The hyperparameters for identification of the three data drift scenario (Periodically, and Abrupt mean shift over time) following algorithm stated in section 3.5 are given in Table 1 later.

It can be noted that apart from the abrupt data drift scenario, in the other two scenarios where mean shifts more smoothly in a periodic or monotonic way, the definition of a data drift itself is very much subjective. Naturally, the drift detection aspect in these two cases is also relative to the definition of drift. In fact, it is quite intuitive that the definition of drift will depend on the variability of performance of the underlying model in two different data regimes. For example, if we consider Figure 5(b) and 5(c), the difference in only in historical window size for anomaly score calculation which reflects quite two different scenarios of data drift. In 5(b), the data drift regions are visibly separated in the two half periods where in 5(c), the data drift is considered more locally as the window length is more and the change pattern in mean is comparatively slow. Now the underlying behavior of the model will determine which definition of drift is more likely.

Similar discussion can be anchored around the monotonic mean change or abrupt mean shift scenario, for which drift onset time periods are illustrated with the parameter values mentioned in table 1.

For data drift identification in the abruptly changing mean scenario, it can be understood that a comparatively lesser window length will not be able to identify the drift scenario and a similar observation was made during the simulation experiment. However, this increases false drift discovery as we observed from Figure 7.

We also experimented with a scenario when there was no data drift and it actually did not detect any drift but the sensitivity is sharp around the historical window size. We could identify no drift situation by increasing the window size, starting from 15.

The proposed HTM-SPRT did work in a mixed way and to a good extent in an expected way. But some of the tuning parameters seems a bit more important than anticipated. For example, setting the null and alternative hypothesis values differently has lesser impact of drift detection than different historical window size. Another notable observation is the possible need of setting a high bar for type II error (smaller value of b). When the cost of retraining applications is comparatively more than the cost of a drift detection failure, we would like to keep such type II error bound so as to limit multiple local drift detection within a short time span.

| Simulation scene | Window length | Bin threshold | p_null | p_alt | a | b | Anomaly limit k (anomaly is beyond k*$\sigma_{roll}$) |
|---|---|---|---|---|---|---|---|
| Shock (Figure 7) | 15 | 0.65 | 0.45 | 0.5 | 0.05 | 0.005 | k= 1 |
| Slow changing mean (Figure 6) | 35 | 0.65 | 0.45 | 0.5 | 0.05 | 0.005 | k= 1 |
| Periodically changing mean (Figure 5) | 25,45 | 0.65 | 0.45 | 0.5 | 0.05 | 0.005 | k= 1 |

**Table 1:** Parameters corresponding to different drift detection scenarios.

Comparison with a few competing approaches:

To conceive some meaningful comparison of the proposed approach with some relevant existing approaches, we have conducted an experiment. Kolmogorov-Smirnov (KS) test, Wasserstein distance and Population Stability Index (PSI) are three popular approaches of comparing two sets of data points with respect to the data distribution. However, to build an online drift detector we need to formulate appropriate execution of the above algorithms. One such formulation, keeping a logical similarity with the proposed approach of HTM and SPRT ensemble, is given below.
1. Start with data from a initial window size [1,w] as the reference dataset. Consider data from the time window [w+1,2w] as the target data set.

2. Test the hypothesis that the reference and the target data set are from same probability distribution with the three algorithms stated above at the time point w+1.
3. Change the reference window to [1,w+1] and the target window [w+2,2w+1] and conduct the same testing as in 2.
4. Continue till the end of the available data horizon, till the reference window is [500-2w,500-w] and the target data widow is [500-w+1,500]

To compare with the proposed approach, we considered w=15, 25 and plot identified drift start time points as a vertical line overlaid with the data (similarly as in the proposed approach).
For testing no drift null hypothesis in step 2 mentioned immediately above, we use a two sided Kolmogorov-Smirnov (KS) test (ks.test() in base R package). For PSI, we compare the PSI score psi** with the Z-score obtained from function psi() from R package PDtoolkit and reject no drift null hypothesis if psi** > Z-score. For Wasserstein distance-based approach, we simulate the 95% percentile point from the distribution of Wasserstein distance computed from 500 iteration. In each iteration, Wasserstein distance were computed from two independent random sample of size 50 drawn from N(0,1) distribution. Note that under no data drift null hypothesis, this 95% percentile point corresponds to a threshold for testing the null hypothesis at level $\alpha = 0.05$.

A sample illustration is given in figure 8 below for the scenario with periodically varying mean.

It can be noticed that for the periodically varying mean scenario, the Kolmogorov-Smirnov (KS) and Wasserstein based approach raises alarmingly high alert of drift detection whereas PSI based approach raises very few drift detection alert. In comparison, our proposed approach is way balanced. For the abrupt mean shift scenario, the performance of the alternatives shows similar behavior of high number of alerts in consecutive time points across a time window. A similar behavior of the competing approach is observed in the slowly moving mean change scenario.
From the above discussions, we can note that the alternative approaches are practically not applicable due to very high false positive alarm whereas the proposed approach shows a much better promise in terms of performance and practical usability.

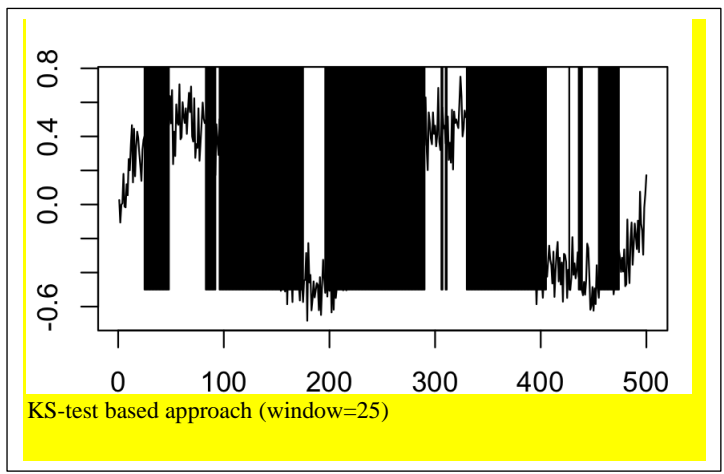

KS-test based approach (window=25)

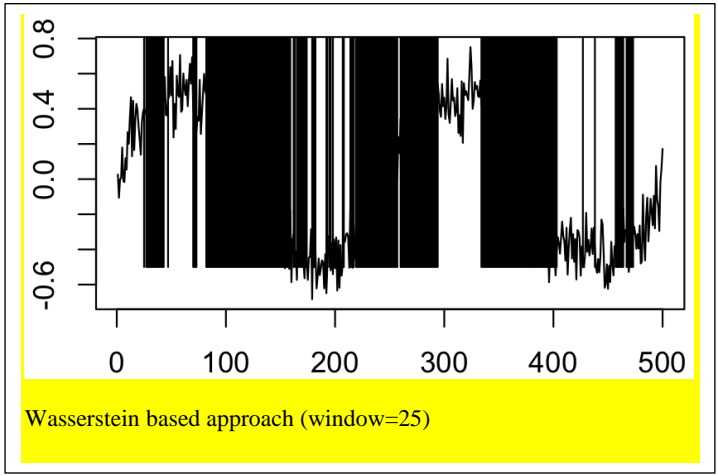
Wasserstein based approach (window=25)

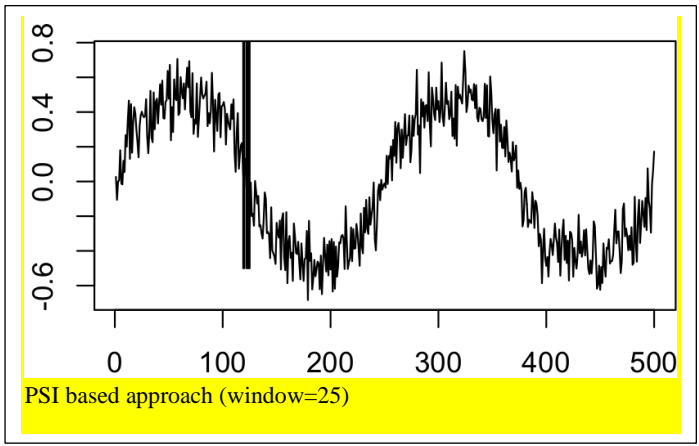
PSI based approach (window=25)

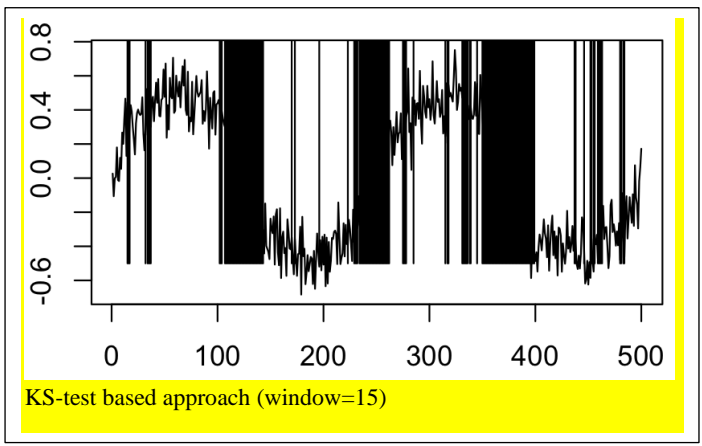
KS-test based approach (window=15)

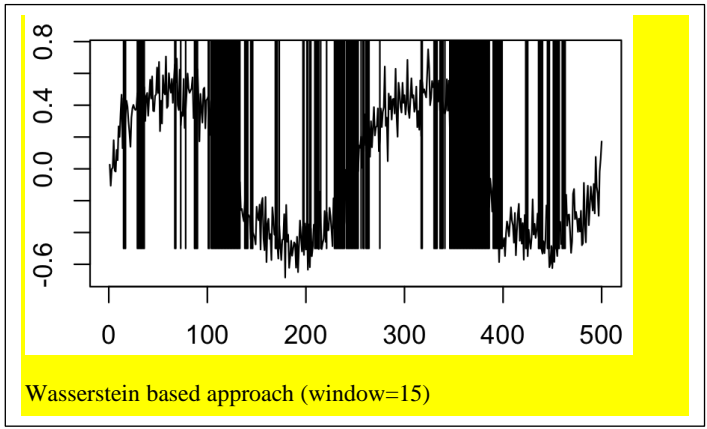

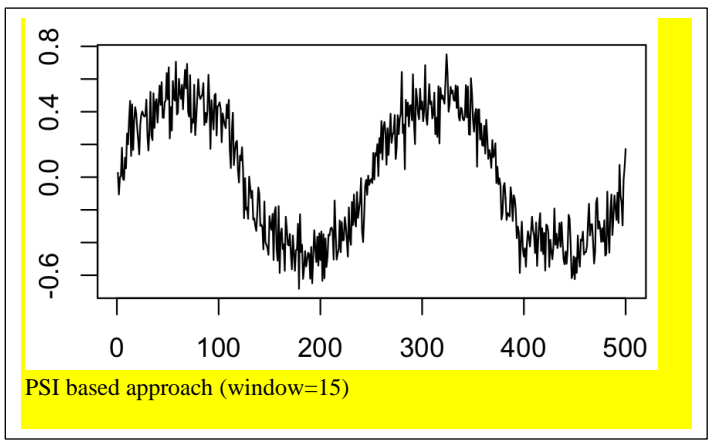

**Figure 8.** Time (X-axis) – Vs – value (Y-axis) Plot from three competing approaches in the periodically varying mean scenario.

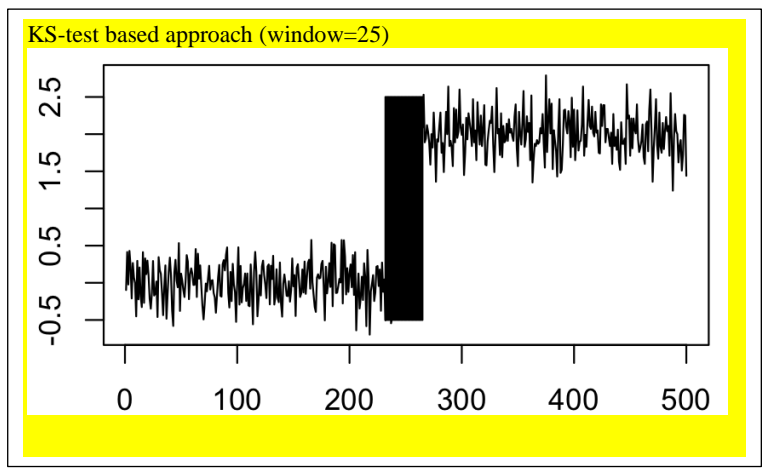

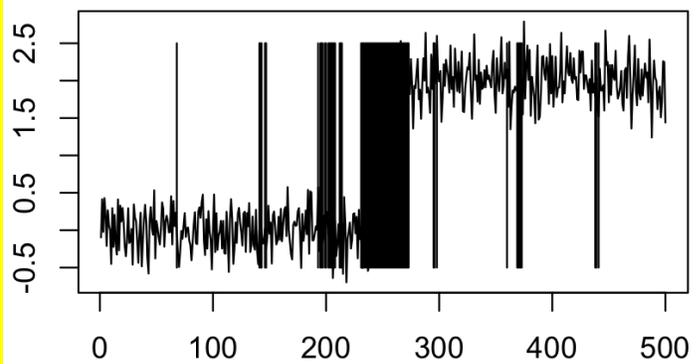

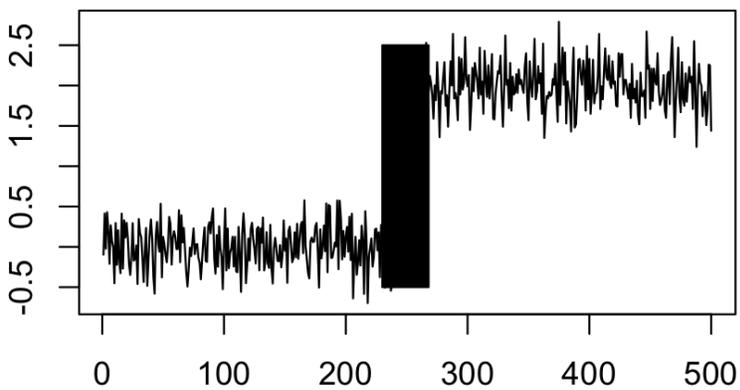

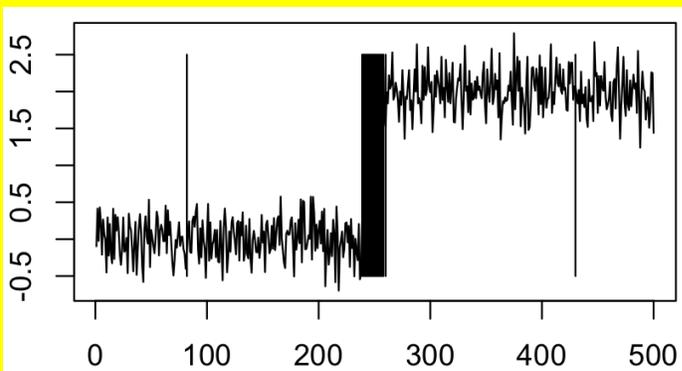

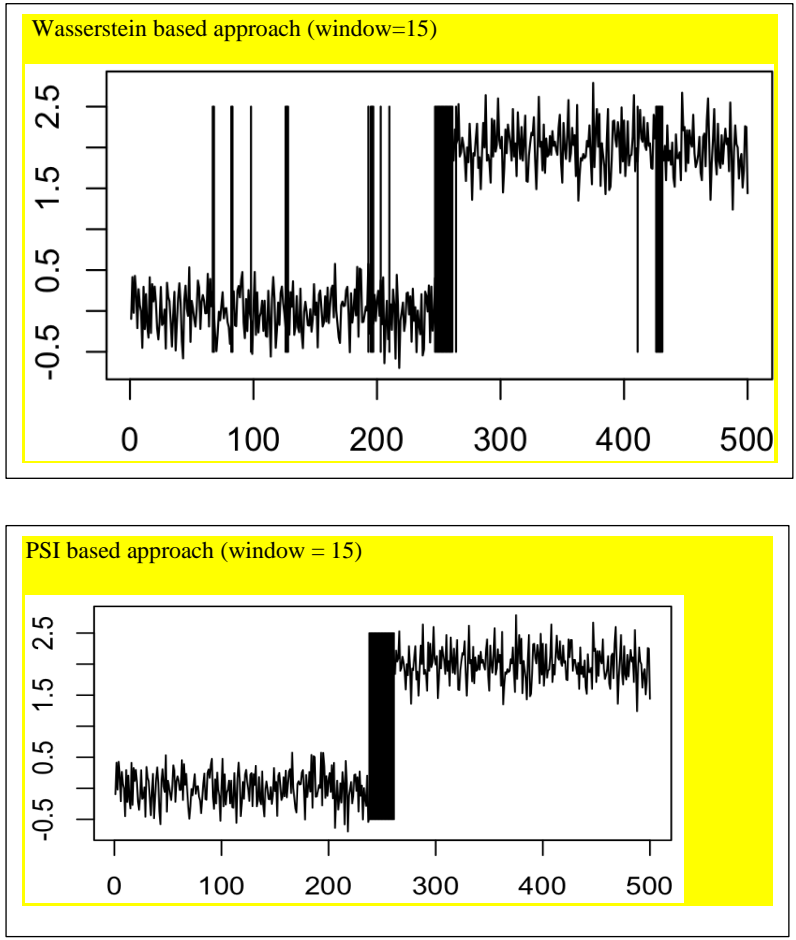

**Figure 9.** Time (X-axis) – Vs – value (Y-axis) Plot from three competing approaches in the abrupt mean shift scenario.

## 4. Data drift detection in supervised scenario: Approach for multivariate data

Currently, HTMs work mainly with univariate. However in the real-world use cases, we may have labelled multidimensional data which current HTM can't handle. To extend HTM application into multivariate data setup, we propose to leverage neural network as explained in the following subsections.

In a supervised scenario when labelled data (e.g., data with outlier tag) is available, we propose an extension of univariate HTM based approached to multivariate setup through a neural network combiner. In this approach we first fit individual HTM model for marginal dimensions. Next, a neural network is trained to classify observations as outlier where the input layer of the network consumes the output of fitted HTM model. For loss computation, label of the original data is utilized.

### 4.1 Application of HTM in the multivariate scenario

As a motivating example, let us consider telecom domain data where different features like latency, throughput and user load etc. from a telecom network are available at cell level (specific locality, for example). These features typically represent different aspect of the network behavior. The anomalous behavior of a network can originate typically from a combination of multiple such aspects. Hence, anomaly from a multidimensional view should be the natural way to detect anomalous behavior of the network at a given time point. Independent feature level view may not correctly identify the anomalous state.

For example, the latency and throughput marginally may not be anomalous but put together, it can detect an unusual scenario where for the specific throughput the observed latency may not be a normal behavior of the network.

To extend the efficient anomaly detection capability of HTM from univariate setup into a multivariate scenario, we propose to utilize a neural network (NN) to combine the univariate HTM output and detect anomaly from a multivariate perspective. In the following subsection we discuss the detailed methodology.

**4.2 Neural network to combine the output of multiple HTM cells for multivariate data**

The inputs and outputs of the neural network are given as follows:

- Corresponding to each data dimension, we fit one HTM cell. Output of HTM cells put together for a specific time point is inputted to the NN as one data point with same time label.

  Note that data other than HTM output can be combined directly with the HTM output form the input data to the NN.

- The output of the neural network is a binary number representing whether the combination of inputs at a given time represents an actual anomaly. We train the neural network using the ground truth labelled (normal/anomalous) data.

- A trained neural network can predict anomalous observation from test data based on similar input features mentioned above

**4.3 Anomaly detection using existing models as the ground truth of the anomalies**

The primary challenge was to create outlier tag as limited telecom domain expert support was available. We did handle this by applying the multiple outlier detection (OD) algorithm on the data and tagging an observation as outlier if 80% agreement is observed from the multiple algorithms. We used PyOD package (Github, PyOD) for this benchmarking where twelve OD algorithms were chosen suitably.

The data preprocessing however was a lengthier step where linear, first and second order interaction was considered for feature generation. Subsequent PCA for dimension reduction was executed. In the end we worked with 54 PCA dimensions where the initial data was of 295 dimensions.

A brief overview of these algorithms from PyOD package that were used for anomaly benchmarking are given below.
- Linear Models for Outlier Detection: This includes the following.
    - PCA: Principal Component Analysis use the sum of weighted projected distances to the eigenvector hyperplane as the outlier scores)

- MCD: Minimum Covariance Determinant (uses the Mahalanobis distances as the outlier scores)
- OCSVM: One-Class Support Vector Machines
- Proximity-Based Outlier Detection Models: This includes the following.
  - LOF: Local Outlier Factor
  - CBLOF: Clustering-Based Local Outlier Factor
  - kNN: k Nearest Neighbors (use the distance to the kth nearest neighbor as the outlier score)
  - Median kNN Outlier Detection (use the median distance to k nearest neighbors as the outlier score)
  - HBOS: Histogram-based Outlier Score
- Probabilistic Models for Outlier Detection:
  - ABOD: Angle-Based Outlier Detection
- Outlier Ensembles and Combination Frameworks: This includes the following.
  - Isolation Forest
  - Feature Bagging
  - LSCP

### 4.4 Application of HTMs to identify anomalies in a sample dataset of KPIs

The applications include the following steps.

1. **Generating HTM output:** Once the 54 columns had been generated, we used HTM on each of the columns to identify the anomalies in each. We treated each of the columns as a HTM input stream. After applying the HTM algorithm to identify the anomalies in each column, we resulted in a list of anomalies (0 or 1).
2. **Training the Neural Network (NN)**: We then trained a neural network of two hidden layers, one input and one output layer and optimized it with Keras Optimizer with 120 as epoch as the stopping rule. The HTM outputs of the 54 columns was the input to the neural network. The output was a binary value or either 0 (indicating absence of anomaly) or 1 (denoting presence of an anomaly).
3. **Performance of the NN combiner**: The NN combiner did a decent job of 90% outlier detection.

### 5. Conclusion

In this paper we have explored approaches to use HTMs in the unsupervised and supervised scenarios for anomaly detection in combination with SPRT and Neural network respectively. For the supervised scenario, we used a method to transform univariate HTM based anomaly detection approach to multivariate anomaly detection via combining marginal HTM anomaly detection probability through a multilayered neural network. The approach is experimented on a sample multivariate time series data consisting of KPIs. This simple combination shows reasonable accuracy in detecting anomalous events.

For the unsupervised scenario, we have used HTMs in combination with a statistical technique known as SPRT to detect data drift. The advantage of an HTM based approach lies in the instantaneous use where expensive training phase is not necessary, and anomaly detection can efficiently take place on streaming/real time/near time data scenario which is specifically suitable for many real-life use cases.

We have observed that the proposed approach is capable to identify data drift in a streaming data with strongly significant lower number of false alarm than existing approaches based on familiar population homogeneity test (Kolmogorov-Smirnov (KS) test, Wasserstein distance and Population Stability Index based approaches).

We can apply the proposed method as standalone application as well as a part of ML pipeline. Inclusion of this approach in a ML pipeline can make the ML application more efficient by raising alarm of a data drift start period and data drift end period. It can be noted that after a data drift start is identified, the data drift end period can be identified as the next data drift start signal.

**Limitations:** The proposed approach is dependent on multiple hyperparameter specifications which are either to be supplied as a derivative of domain knowledge, or to be estimated through elaborate simulation based on enough historical data. Efficient choice of hyperparameters is important to balance a strike between recall and precision. Though we have proposed a few basic parameter combinations from a primary simulation experiment, this should come ideally as a part of tuning stage. For supervised scenario, the objective function is quite trivial but in the unsupervised scenario, specifically in a multidimensional set up, we need to pursue a deeper study to understand data driven optimality criteria and hence set a proper data driven hyper-parameter tuning.

**Future Scope:** There are couple of gaps that we would like to fill in further continuation of this study as mentioned below.

*Hyper-parameter tuning:* A possible direction may be to consider relative population stability in different drifted data regime while identifying most unstable transition phase as indicator of data drift. In one dimensional scenario, an early experiment with data variance as a measure of stability seems to work reasonably. Corresponding multivariate extension is needed to be studied in detail.

*Multivariate extension:* Multivariate extension of HTM from first principal and combine it with existing multivariate SPRT is another interesting direction to explore.

*Composite Hypothesis in SPRT:* It can be noted that data drift detection is mapped with simple hypothesis construction via HTM and testing via SPRT. Ideally, we should consider both null and alternative hypotheses in composite status through ranges of the parameter in Bernoulli distribution corresponding to a drifted or not drifted data scenario. Hence the right variation of SPRT can be explored for the testing exercise.

**Conflict of Interest**
The authors confirm that there is no conflict of interest to declare for this publication.

## Acknowledgments

This research did not receive any specific grant from funding agencies in the public, commercial, or not-for-profit sectors. The authors would like to thank the editor and anonymous reviewers for their comments that help improve the quality of this work.